\documentclass{article}
\usepackage{ijcai24}

\usepackage{times}
\usepackage{soul}
\usepackage{url}
\usepackage[hidelinks]{hyperref}
\usepackage[utf8]{inputenc}
\usepackage[small]{caption}
\usepackage{graphicx}
\usepackage{amsmath}
\usepackage{amsthm}
\usepackage{booktabs}
\usepackage{algorithm}
\usepackage[switch]{lineno}
\usepackage{algpseudocode}

\usepackage{amssymb}
\usepackage{titlesec}
\usepackage{float}
\usepackage[table,xcdraw]{xcolor}
\titleformat{\paragraph}
{\normalfont\normalsize\bfseries}{\theparagraph}{1em}{}

\usepackage{lscape}
\usepackage[figuresright]{rotating}
\usepackage{tabularx}
\usepackage{booktabs}

\usepackage{graphicx}
\usepackage{physics}
\usepackage{amsmath}
\usepackage{tikz}
\usepackage{mathdots}
\usepackage{yhmath}
\usepackage{cancel}
\usepackage{color}
\usepackage{siunitx}
\usepackage{array}
\usepackage{multirow}
\usepackage{amssymb}
\usepackage{gensymb}
\usepackage{tabularx}
\usepackage{extarrows}
\usepackage{booktabs}
\usetikzlibrary{fadings}
\usetikzlibrary{patterns}
\usetikzlibrary{shadows.blur}
\usetikzlibrary{shapes}
\usepackage{diagbox}
\usepackage{bbding}
\usepackage{amsthm}

\theoremstyle{definition}

\newcolumntype{C}{>{\centering\arraybackslash}X}

\title{Deep Generative Domain Adaptation with Temporal Relation Knowledge for Cross-User Activity Recognition}

\author{
Xiaozhou Ye$^1$
\and
Kevin I-Kai Wang$^2$\\
\affiliations
$^1$ $^2$Department of Electrical, Computer, and Software Engineering, The University of Auckland, Auckland, New Zealand\\
\emails
xye685@aucklanduni.ac.nz,
kevin.wang@auckland.ac.nz
}

\begin{document}

\maketitle

\begin{abstract}
In human activity recognition (HAR), the assumption that training and testing data are independent and identically distributed (i.i.d.) often fails, particularly in cross-user scenarios where data distributions vary significantly. This discrepancy highlights the limitations of conventional domain adaptation methods in HAR, which typically overlook the inherent temporal relations in time-series data. To bridge this gap, our study introduces a Conditional Variational Autoencoder with Universal Sequence Mapping (CVAE-USM) approach, which addresses the unique challenges of time-series domain adaptation in HAR by relaxing the i.i.d. assumption and leveraging temporal relations to align data distributions effectively across different users. This method combines the strengths of Variational Autoencoder (VAE) and Universal Sequence Mapping (USM) to capture and utilize common temporal patterns between users for improved activity recognition. Our results, evaluated on two public HAR datasets (OPPT and PAMAP2), demonstrate that CVAE-USM outperforms existing state-of-the-art methods, offering a more accurate and generalizable solution for cross-user activity recognition.
\end{abstract}

\section{Introduction}

Human Activity Recognition (HAR) is an essential field within Human-Computer Interaction, ubiquitous computing \cite{ma2019attnsense}, and the Internet of Things \cite{abdallah2018activity}. It involves identifying human activities using sensor data and contextual information \cite{qian2019novel}. HAR finds applications in diverse areas like medical treatment, assisted living, fitness, security, and home automation. Current HAR methods, especially those processing time-series sensor data, operate under the assumption that training and testing data are drawn from the same distribution, meaning they are independent and identically distributed (i.i.d.) \cite{wilson2020survey}. This approach assumes that a model trained on source data will perform similarly on target data, as long as both sets of data are from the same domain with consistent features and distribution characteristics. However, this is often not the case in real-world scenarios, where training and testing datasets may have different distributions due to data heterogeneity, also known as the out-of-distribution (o.o.d.) problem. This discrepancy can lead to reduced model performance when applied to new, unseen data \cite{patel2015visual}.

Transfer learning, specifically domain adaptation \cite{pan2009survey}, is a method that addresses this data heterogeneity issue. Its core principle is to identify and leverage common knowledge to minimize differences in data distribution between the source and target domains. Current research in domain adaptation primarily focuses on static data, like images, where each sample is considered i.i.d. within its domain \cite{patel2015visual}. This assumption is extended to time-series data in HAR, treating each data segment as i.i.d. in its domain \cite{soleimani2021cross}. However, this approach overlooks the inherent temporal relations in time-series data, where consecutive data segments are often interdependent. For example, in activity recognition, activities like walking consist of interconnected sub-activities (e.g., lifting the leg, moving it forward, placing the foot down), each with temporal dependencies. The prevalent oversight in current research methods is the insufficient attention to these temporal dependencies, which are crucial for understanding and accurately recognizing human activities. This lack of focus results in models that fail to capture the common sequential dynamics of activity data, leading to suboptimal performance on domain adaptation, especially when applied across diverse users.

This paper introduces a novel approach called Conditional Variational Autoencoder with Universal Sequence Mapping (CVAE-USM) for time series domain adaptation in HAR. This method focuses on capturing temporal relations in activity data by USM encoding, enabling better adaptation between different users' data. By preserving these temporal relations, our approach effectively regularizes the adaptation process, enhancing the alignment of sub-activity distributions across users. This results in improved performance in domain adaptation tasks. Moreover, We propose a new generative model of variational autoencoder (VAE) based architecture that captures temporal relation knowledge for cross-user HAR, which enhances the generalized capability of VAE generative model. Our extensive testing on two public HAR datasets shows that this method outperforms existing approaches in time series domain adaptation, particularly in cross-user scenarios. 

The paper is organized as follows: Section 2 reviews related work in HAR, transfer learning, and domain adaptation. Section 3 details our proposed CVAE-USM method, focusing on capturing and aligning temporal relations. Section 4 outlines our experimental setup and compares our method against existing approaches. Finally, Section 5 concludes the paper and suggests directions for future research.

\section{Related work}

\subsection{Cross-user human activity recognition}

Human Activity Recognition (HAR) is a fundamental aspect of ubiquitous computing, playing a vital role in enhancing daily human activities. Its primary goal is to recognize and analyze human behaviors by interpreting high-level knowledge derived from multi-modal sensor data and contextual information. Based on sensor modalities, HAR can be categorized into five distinct types: Smartphones/wearable sensors-based HAR, ambient sensors-based HAR, device-free sensors-based HAR, vision-based sensors-based HAR, and other modality sensors-based HAR \cite{chen2021deep}\cite{lentzas2020non}. This paper specifically delves into HAR using wearable sensors.

Within machine learning research, sensor-based HAR is often approached as a time series classification problem \cite{andreas2014tutorial}. Various classification models, including ensemble learning \cite{sekiguchi2020ensemble}, SVM \cite{bulling2012multimodal}, and HMM \cite{amft2005detection}, have been proposed to solve HAR challenges. With advancements in deep learning, numerous state-of-the-art results have been achieved for a wide range of tasks. Deep learning-based HAR techniques excel in learning high-level features and autonomously extracting features from extensive data sets \cite{qian2019novel}. However, these approaches predominantly rely on the assumption that the training and testing data are drawn from the same distribution, meaning the data is considered independent and identically distributed (i.i.d.) \cite{wilson2020survey}. This assumption often fails in real-world applications, where the collected training and testing datasets are out-of-distribution (o.o.d.). Our focus in this paper is on addressing the challenges associated with sensor-based HAR in the context of o.o.d. data.

There are several categories of sensor-based HAR o.o.d. challenges: First, differences in data may result from the use of various sensor types, platforms, manufacturers, and modalities, leading to diverse data formats and distributions \cite{xing2018enabling}. Second, the data pattern might change over time, a phenomenon known as concept drift \cite{lu2018learning}. For example, a person's walking pattern could be influenced by changes in health status. Third, behavioral differences between individuals can be significant \cite{xiaozhou2023temporaloptimal}; for instance, walking pace may vary from person to person. Fourth, the positioning of physical sensors on the body \cite{rokni2018autonomous} or the layout of environmental sensors in smart homes \cite{sukhija2019supervised} can also lead to different data distributions. Our research specifically targets sensor-based HAR o.o.d. problems stemming from behavioral differences among individuals.

\subsection{Transfer learning and domain adaptation}

Transfer learning enables models to learn from one or more source domains and then apply this knowledge to related target domains lacking labelled data. This process is crucial to address the out-of-distribution (o.o.d.) problem by minimizing distribution differences between the source and target domains. Domain adaptation, a subset of transfer learning, specifically tackles the o.o.d. issue while maintaining the same task across source and target domains. This approach typically involves leveraging labelled data from the source domain and unlabeled data from the target domain \cite{wilson2020survey}.

Domain adaptation has experienced significant growth, particularly in the area of feature-based transfer learning \cite{pan2009survey}. Techniques like Subspace Alignment (SA) \cite{fernando2013unsupervised} focus on finding similarities between the feature subspaces of the source and target domains, using principal component analysis and linear transformations. Optimal Transport for Domain Adaptation (OTDA) \cite{flamary2016optimal}, is grounded in optimal transport theory, aiming to establish a cost-effective correspondence between different domains. Deep domain adaptation has also seen advancements with approaches like the Domain Adversarial Neural Network (DANN) \cite{ganin2016domain}, which employs adversarial learning to train a model so that its features cannot be used to distinguish between source and target domains. This promotes the generation of domain-invariant features. FedMAT \cite{shen2022federated} works by treating each individual's data as a distinct task within a federated learning system. It uniquely combines a central shared feature representation network with individual-specific networks equipped with attention modules in decentralized nodes, enabling the learning of both shared and individual-specific features from multi-modal sensor data.

These feature-based methods primarily target static data like images, often applying the same framework to time series data \cite{lu2021cross}\cite{li2020simultaneous}. However, this approach may be less effective for time series data, where temporal relationships are crucial. Current domain adaptation strategies often overlook these temporal relations, leading to models that may not perform reliably, especially in cross-user HAR applications where models trained on data from one user are applied to others. Incorporating temporal relation knowledge could enhance the identification of commonalities between users, improving domain adaptation in sensor-based HAR. This paper focuses on exploring and integrating temporal relation knowledge to enhance domain adaptation effectiveness in cross-user HAR.

\section{Method}
\subsection{Problem formulation}
In a cross-user HAR problem, a labelled source user $\displaystyle S^{Source} =\left\{\left( x_{i}^{Source} ,\ y_{i}^{Source}\right)\right\}_{i=1}^{n^{Source}} $ drawn from a joint probability distribution $\displaystyle P^{Source} $ and a target user $\displaystyle S^{Target} =\left\{\left( x_{i}^{Target} ,\ y_{i}^{Target}\right)\right\}_{i=1}^{n^{Target}} $ drawn from a joint probability distribution $\displaystyle P^{Target} $, where $\displaystyle {n^{Source}} $ and $\displaystyle {n^{Target}} $ are the number of source and target samples respectively.  $\displaystyle S^{Source}$ and $\displaystyle S^{Target}$ have the same feature spaces (i.e. the set of features that describes the data from sensor readings) and label spaces (i.e. the set of activity classes). The source and target users have different distributions, i.e., $\displaystyle P^{Source} \neq P^{Target} $, which means that even for the same activity, the sensor readings look different between the two users. Given the labelled source user data and unlabelled target user data, the goal is to obtain the labels $\displaystyle \left\{\left( y_{i}^{Target}\right)\right\}_{i=1}^{n^{Target}} $ for the target user activities.

\subsection{Conditional Variational Autoencoder with Universal Sequence Mapping}

In this study, we present the Conditional Variational Autoencoder with Universal Sequence Mapping (CVAE-USM), a novel approach for cross-user HAR. The foundation of this method is the understanding that human physical movements are influenced by preceding actions, and these temporal dependencies are generally consistent across different individuals. By capturing these shared temporal patterns in human activity time series data, our model aims to enhance its performance and generalizability for cross-user HAR tasks.

\begin{figure}[h!]
\centering
\scalebox{1.0}{
  \includegraphics[width=\columnwidth]{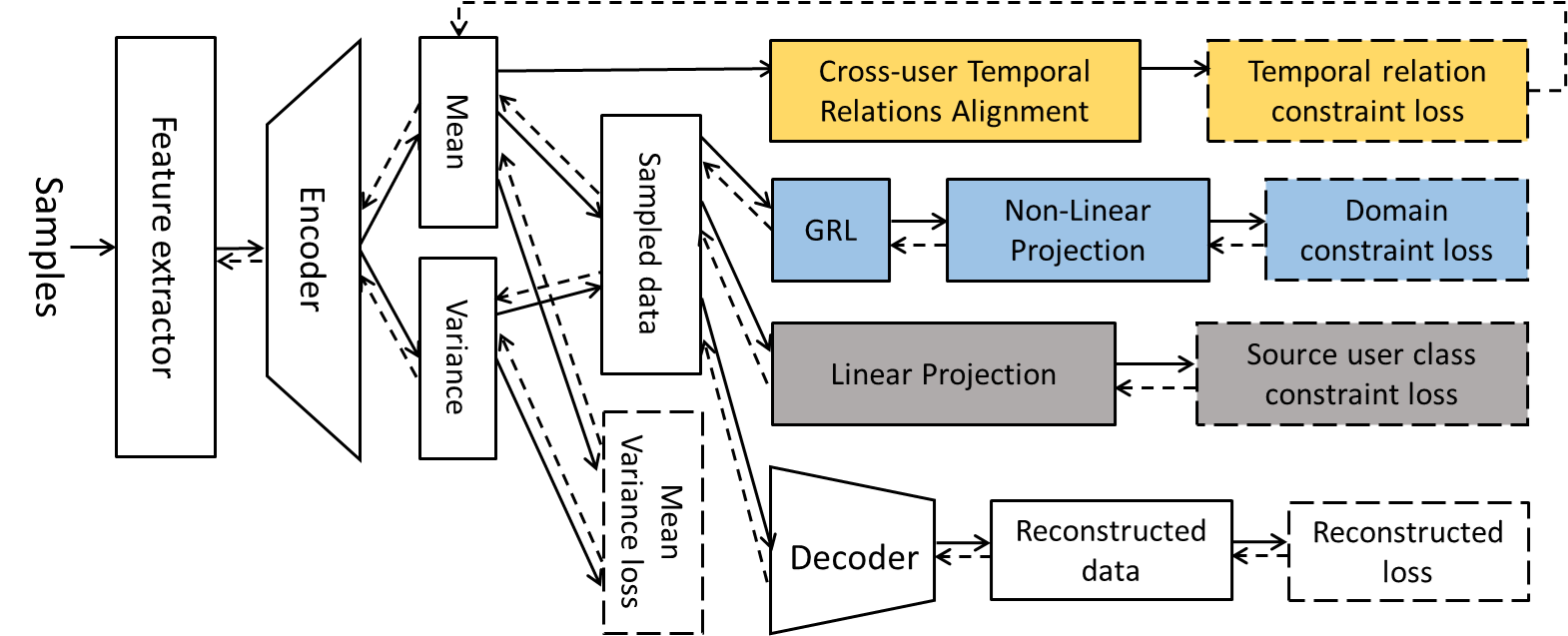}
}
\caption{CVAE-USM Learning Process.\label{CVAE_USM_framework}}
\end{figure}

\subsubsection{Generative Model Architecture}

Our approach utilizes a generative model framework, as depicted in the CVAE-USM Learning Process shown in Figure~\ref{CVAE_USM_framework}, to achieve effective model generalization. Generative models are advantageous in that they can produce new data samples that are similar to the training data, aiding in data augmentation and improving generalization capabilities. Even in scenarios where some class labels are missing during training, generative models focus on modelling the data distribution itself, which is more effective in uncovering the underlying data patterns and structures \cite{ruthotto2021introduction}. For cross-user HAR, where labels for target users are not available (or only partially available), generative models are particularly useful as they are capable of modelling user-invariant activity data distributions. They achieve this by encoding data into a compact latent space, capturing the underlying factors or patterns in the data \cite{van2010activity}.

In implementing our method, we employ the Variational Autoencoder (VAE) architecture \cite{Kingma2014} as the foundational generative model. The VAE aims to learn a structured latent space where similar data samples are closely aligned, such as patterns found in time series data, typically following a Gaussian distribution. In the VAE structure, input data (like sensor features in HAR) is compressed into a lower-dimensional latent space via an encoder. This latent space is represented probabilistically with mean and variance parameters, encapsulating intrinsic characteristics of human activities. The VAE employs the reparameterization trick for differentiability during training, sampling from the latent space based on Gaussian distributions and then using a decoder to reconstruct the original input. This method enables the capturing of complex temporal activity patterns, making the VAE an ideal model for cross-user HAR tasks.

The VAE's sampling process offers several benefits for generalization and domain adaptation. The randomness introduced through sampling acts as a regularization form, preventing overfitting and enhancing generalization. By encoding data into a distribution instead of a deterministic decision boundry and encouraging the latent variables to approximate a Gaussian distribution (via the Kullback–Leibler divergence), the latent space becomes more structured and continuous  \cite{rybkin2021simple}. In such a structured latent space, alignment techniques like Gradient Reversal Layer (GRL) adversarial learning, which we will discuss later, can be more effectively applied for domain adaptation in HAR.

Moreover, we implement an adversarial learning strategy, specifically utilizing the Gradient Reversal Layer (GRL) technique \cite{ganin2016domain} as the blue parts in  Figure~\ref{CVAE_USM_framework}. This strategy is integral to modifying the model's loss functions related to domain constraints (i.e. confusing the data distributions between source and target users). The GRL adversarial learning operates by reversing the gradients during the training phase. The primary goal of this technique is to ensure that the features learned by the model are not solely tied to the specific user but also lean towards a generalized model. Through this approach, our model can generate user-invariant representations. Therefore, this strategy is integrated as a constraint condition to the original CVAE for model learning. 

Furthermore, after acquiring the sampled features, they are input into a process focusing on the source user's activity class constraint as the grey parts in  Figure~\ref{CVAE_USM_framework}. This involves using a linear transformation to shape how the data distribution is generated. Specifically, this step takes into consideration the activity class labels associated with the source user. This means that the transformation is guided by the types of activities the source user performed, helping to tailor the model's understanding of the data based on these specific activities. Therefore, this source user class constraint is integrated as another constraint condition to the original CVAE for model learning.

\subsubsection{Cross-user Temporal Relations Alignment}

The key element of the CVAE-USM method is the cross-user temporal relations alignment. This CVAE constraint condition is shown in the yellow parts in Figure~\ref{CVAE_USM_framework}. In this component, Gaussian Mixture Model (GMM), Universal Sequence Mapping (USM), and Wasserstein Distance techniques are included. Each component plays a pivotal role in capturing and adapting temporal relation knowledge for domain adaptation. The overall process of cross-user temporal relations alignment is described in Algorithm~\ref{algo1}.

\begin{algorithm}
\caption{Cross-user Temporal Relations Alignment}
\label{algo1}
\begin{algorithmic}
\State \textbf{Input:} Mean values from CVAE encoder for source ($\mu_{s}$) and target ($\mu_{t}$) users, number of components ($K$) in GMM.
\State \textbf{Output:} Aligned temporal relation features, Wasserstein distance ($W$) temporal relation loss.

\State \textbf{Begin Algorithm}
\State \textbf{1. Sub-Activity Identification with GMM:}
\begin{itemize}
  \item Apply GMM to sensor data for source and target users: $GMM_{init}(\{ \mu_{s}, \mu_{t} \}, K)$.
  \item This step is to identify and decompose complex activities into simpler sub-activities.
\end{itemize}

\State \textbf{2. Temporal Relations Encoding with USM:}
\begin{itemize}
  \item Transform sub-activities into feature space using USM: $F_{s} = \mathcal{U}(Seq_s)$, $F_{t} = \mathcal{U}(Seq_t)$.
  \item This step aims to capture and encode temporal relation for activities. Here, $\mathcal{U}$ denotes the USM function, and $Seq_s$, $Seq_t$ are sequences of encoded sub-activities for source and target, respectively, with each value in the sequences encoded into the set $\{0, K-1\}$.
\end{itemize}

\State \textbf{3. User-Specific Temporal Feature Analysis:}
\begin{itemize}
\item Perform separate GMM clustering on the USM-encoded features for source ($F_{s}$) and target ($F_{t}$) users: $GMM_s(F_{s}, K)$ and $GMM_t(F_{t}, K)$.
\item This step is to analyze the variations and similarities between the sub-activities' data distributions across users. Let $P_{s}$ and $P_{t}$ denote the resulting distributions from $GMM_s$ and $GMM_t$ respectively, capturing distinct temporal patterns of each user group.
\end{itemize}

\State \textbf{4. Data Distribution Alignment with Wasserstein Distance:}
\begin{itemize}
\item Minimize the Wasserstein distance between source and target distributions: $W(P_{s}, P_{t}) = \inf_{\gamma \in \Pi(P_{s}, P_{t})} \mathbb{E}{(x, y) \sim \gamma}[|x - y|]$.
\item This step aims to adjust model parameters to align $P{s}$ and $P_{t}$, focusing on common temporal relations.
\end{itemize}
\State \textbf{End Algorithm}
\end{algorithmic}
\end{algorithm}

The first step is the Gaussian Mixture Model (GMM) for sub-activity identification. GMM \cite{zhang2021gaussian} is a probabilistic model for representing normally distributed subpopulations within an overall population. It is defined as:

\begin{equation}
    p(x) = \sum_{k=1}^{K} \pi_k \mathcal{N}(x | \mu_k, \Sigma_k)
\end{equation}
Where  \( x \) is the data point. \( K \) is the number of Gaussian distributions in the mixture. \( \pi_k \) are the mixture weights, with \( \sum_{k=1}^{K} \pi_k = 1 \) and \( \pi_k \geq 0 \). \( \mathcal{N}(x | \mu_k, \Sigma_k) \) denotes the Gaussian distributions with mean \( \mu_k \) and covariance \( \Sigma_k \).

GMM identifies sub-activities hidden in the sensor data, effectively decomposing complex activities into simpler sub-activities which represent common knowledge across different users. This model identifies inherent sub-structures within the dataset, irrespective of their originating domain (source or target). This decomposition is vital for two reasons: firstly, it simplifies the complex temporal patterns into more manageable segments, and secondly, the common sub-activities are used for subsequent temporal relation capture and analysis between the source and target domains.

Following the identification of sub-activities, the next step is using USM for temporal feature encoding. This step is crucial for transforming the identified sub-activities into a feature space that encapsulates both their spatial structural and temporal characteristics. USM \cite{almeida2002universal} is a technique for encoding sequences to capture their inherent structure and temporal relations. It can be mathematically represented as:

\begin{equation}
    F = \mathcal{U}(Seq)
\end{equation}
Where \( F \) represents the feature space encoding the sequence. \( \mathcal{U} \) is the USM function. \( Seq \) is the sequence of sub-activities.

USM plays a pivotal role in maintaining the integrity of temporal relations within the activity sequences. This encoding not only represents the sequence of sub-activities but also preserves the temporal dynamics associated with each activity – that is, the temporal order and sequence of sub-activities, particularly when analyzing complex human behaviours, thus offering a more precise representation of human activities. This encoding respects the chronological order of events, ensuring that the temporal dynamics—such as the duration, sequence, and intervals between sub-activities—are accurately captured and reflected in the feature vectors. This is particularly important for understanding how sub-activities unfold over time. 

By effectively encoding the temporal order, USM enables the identification of recurrent temporal patterns and relationships within the data. This can include regular sequences of activities, the typical duration of certain activities, or the common intervals between sub-activities. Moreover, USM-encoded features facilitate the comparison of temporal sequences across different users. By providing a uniform method of encoding, USM allows for the direct comparison of temporal features, supporting the data distribution alignment of temporal relations across users. This step paves the way for the following temporal relation alignment across users.

The next step is user-specific feature analysis. The USM-encoded features, now rich with temporal relational information, undergo separate GMM clustering for the source and target domains. This bifurcation is key to understanding how temporal patterns manifest uniquely in each domain. By analyzing these user-specific GMMs, the approach gains insights into the variations and similarities in activity patterns across users. This comparative analysis is essential for identifying user-specific differences in temporal activity structures.

Finally, Wasserstein Distance is applied for data distribution alignment across users with the above-learned temporal relation knowledge. The Wasserstein Distance \cite{wu2020domain}, also known as the Earth Mover's Distance, measures the distance between two probability distributions. It is defined as:
\begin{equation}
    W(P_s, P_t) = \inf_{\gamma \in \Pi(P_s, P_t)} \mathbb{E}_{(x, y) \sim \gamma}[\|x - y\|]
\end{equation}
Where \( W(P_s, P_t) \) is the Wasserstein Distance between two distributions. \( P_s \) and \( P_t \) are the probability distributions of source and target users being compared. \( \gamma \) represents a joint distribution with marginals \( P_s \) and \( P_t \). \( \Pi(P_s, P_t) \) denotes the set of all possible joint distributions \( \gamma(x, y) \) with marginals \( P_s \) and \( P_t \). \( \mathbb{E}_{(x, y) \sim \gamma}[\|x - y\|] \) is the expected value of the Euclidean distance between sampled points \( x \) and \( y \) from \( \gamma \).

Wasserstein distance is computed using the means of the GMMs from the source and target users. This distance metric quantifies the disparity in the temporal feature distributions between the two domains. Minimizing the Wasserstein distance is used to align the source and target user distributions in the temporal relation view based on the idea that source and target users should have common temporal relation sub-activities. This alignment is central to the model's ability to generalize and accurately recognize activities across users.

\section{Experiments}

\subsection{Datasets and Experimental setup}

For validating our cross-user domain adaptation method in sensor-based HAR, we employed two widely recognized public datasets as outlined in Table~\ref{tab_datasets_info}. Our focus was on the practical application scenario involving a smartwatch. To this end, we only utilized sensor data from accelerometers and gyroscopes positioned on the right lower arm. Below, we provide a brief overview of each dataset, highlighting their key aspects.

\begin{table}[h!]
\caption{Two sensor-based HAR datasets information.}
\label{tab_datasets_info}
\centering
\resizebox{\columnwidth}{!}{%
\begin{tabular}{|p{1.2cm}|p{1.2cm}|p{1.5cm}|p{7.0cm}|}
\hline
\textbf{Dataset} & \textbf{Subjects} & \textbf{\#Activities} & \textbf{Common Activities} \\ \hline
OPPT & S1, S2, S3 & 4 & 1 standing, 2 walking, 3 sitting, 4 lying \\ \hline
PAMAP2 & 1, 5, 6 & 11 & \begin{tabular}[c]{@{}l@{}}1 lying, 2 sitting, 3 standing, 4 walking, \\ 5 running, 6 cycling, 7 Nordic walking, \\ 8 ascending stairs, 9 descending stairs, \\ 10 vacuum cleaning, 11 ironing\end{tabular} \\ \hline
\end{tabular}}
\end{table}

The OPPORTUNITY (OPPT) dataset \cite{chavarriaga2013opportunity} contains recordings of people doing morning activities in a natural way, without strict guidance, recorded at 30 Hz. The Physical Activity Monitoring (PAMAP2) dataset \cite{reiss2012introducing} features over 10 hours of data recorded at 100 Hz, with subjects performing specific, pre-defined activities.

In the experimental setup for cross-user HAR, the sliding window technique is employed for data segmentation. This widely-used approach in sensor-based HAR involves creating fixed time intervals of 3 seconds with a 50\% overlap between windows, in line with standard practices in HAR tasks \cite{wang2018impact}. The study compares four methods, divided into traditional domain adaptation and deep domain adaptation categories. Notably, the TrC method requires a few target domain labels for fine-tuning, unlike other methods which do not require any labelling in the target domain.

\textbf{Traditional Domain Adaptation Methods:}

\textbf{CORAL} (Covariance Alignment) \cite{sun2016return}: Aims to align the covariance of feature layers across domains, fostering domain-agnostic features.

\textbf{SOT} (Substructure Optimal Transport) \cite{lu2021cross}: Focuses on underlying domain structures to facilitate detailed substructure mapping, balancing broad and specific mapping.

\textbf{Deep Domain Adaptation Methods:}

\textbf{DANN} (Domain-Adversarial Neural Network) \cite{ganin2016domain}: A deep learning approach using adversarial training to produce features indistinguishable between domains by a discriminator.

\textbf{TrC} (Temporal Regularized CNNs) \cite{rokni2018personalized}: Employs CNNs for discerning temporal features, trained on source domain data and fine-tuned on the target domain.

Hyper-parameters tuning is implemented on these methods with the methodology suggested by Flamary et al. \cite{flamary2016optimal}, specifically to prevent overfitting during testing. We divide the target user's data into two parts: a validation subset and a test subset. The validation subset is utilized to fine-tune the hyper-parameters to achieve the highest accuracy. After optimizing these hyper-parameters, we evaluate the performance of the model on the test subset. The key metric we use for evaluation is the classification accuracy specific to the target user.

\subsection{Cross-user HAR performance results}

We evaluate the performance of various methods on the OPPT and PAMAP2 datasets. Each method is tested for its ability to transition between individual users. Within each dataset, three users are selected randomly, and a one-to-one cross-user HAR task is conducted for every possible user pair.

\begin{figure}[h!]
\centering
\includegraphics[width=\columnwidth]{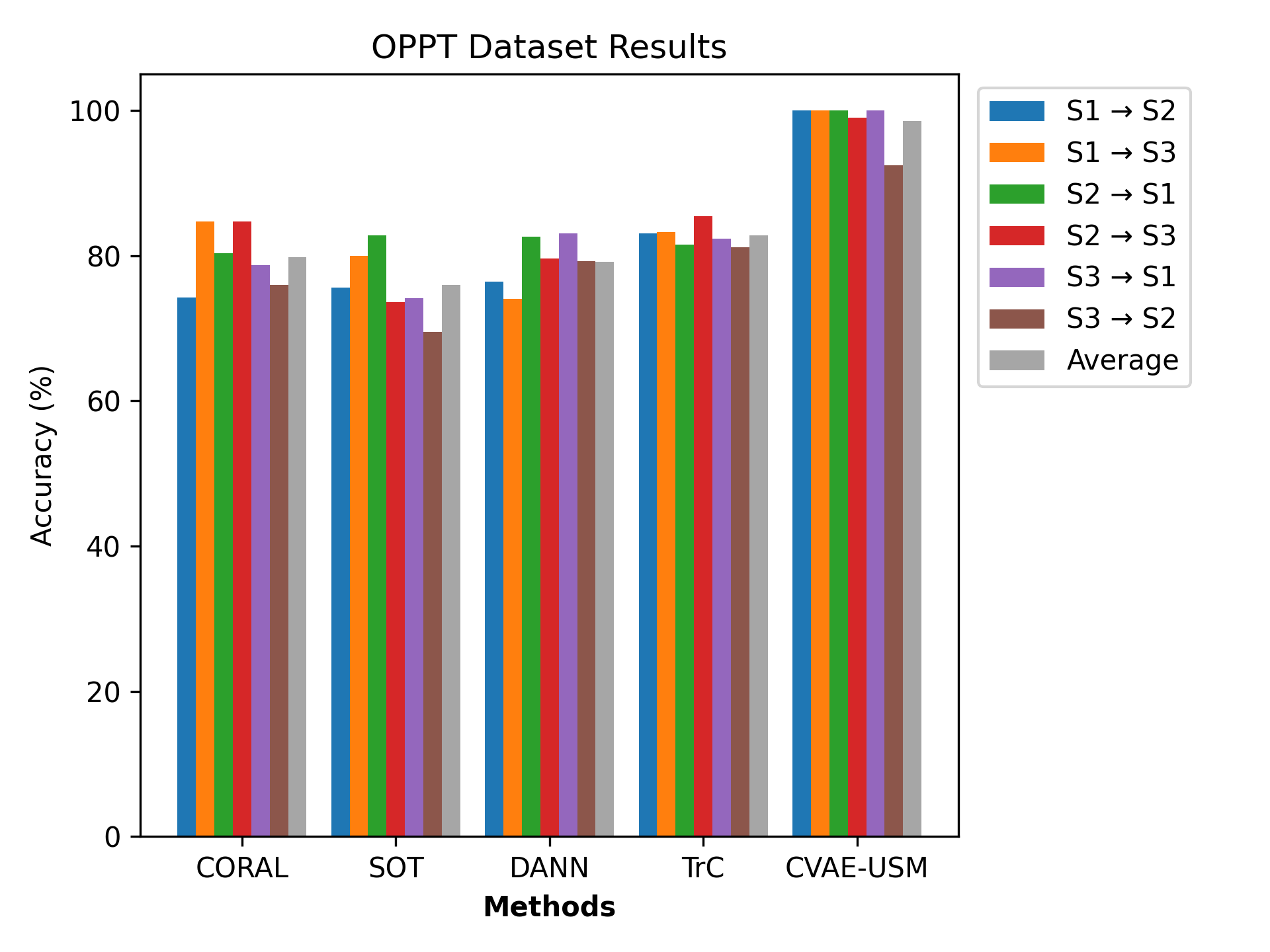}
\caption{OPPT dataset classification results.\label{OPPT_Dataset_Results}}
\end{figure}

In the results of OPPT dataset as shown in Figure~\ref{OPPT_Dataset_Results}, CVAE-USM stands out by achieving almost 100\% accuracy in all testing scenarios. This exceptional performance underscores CVAE-USM's adaptability and effectiveness in various conditions specific to the OPPT dataset. TrC also achieves a reasonably good performance, consistently reaching accuracy levels in the low to mid-80\% range. This indicates its competence in the OPPT dataset, though it doesn't match the superior performance of CVAE-USM. Another method DANN demonstrates moderate performance, with accuracy generally fluctuating between the mid-70\% and mid-80\% range. SOT and CORAL exhibit slightly lower performance levels compared to the others. One notable distinction contributing to the performance discrepancies is the utilization of temporal information. CVAE-USM effectively leverages temporal data, contributing significantly to its superior accuracy. In contrast, TrC considers the temporal features , however it does not exploit temporal information as effectively as CVAE-USM, which partially explains its lower performance metrics.

\begin{figure}[h!]
\centering
\includegraphics[width=\columnwidth]{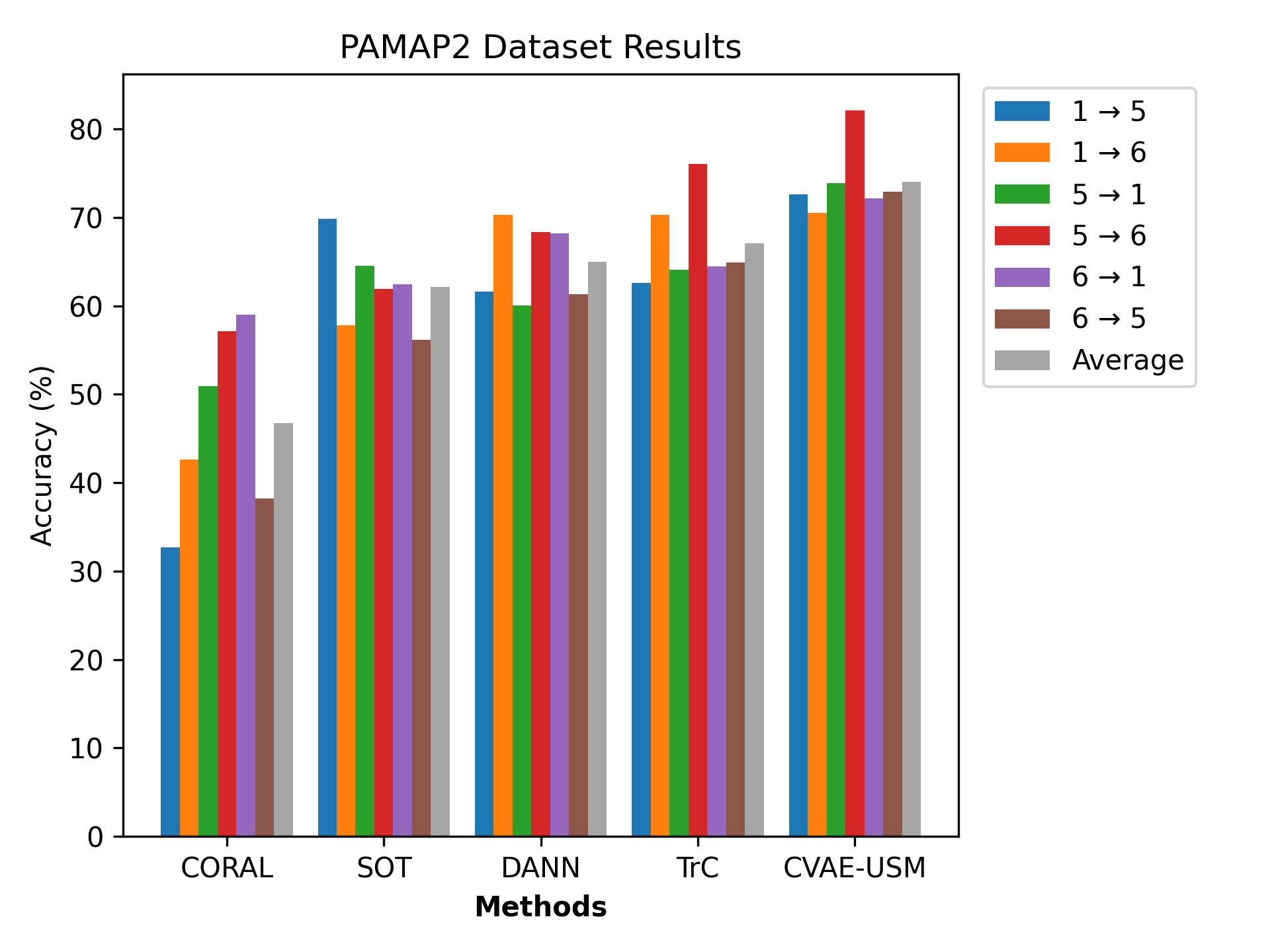}
\caption{PAMAP2 dataset classification results.\label{PAMAP2_Dataset_Results}}
\end{figure}

In the results of PAMAP2 dataset as shown in Figure~\ref{PAMAP2_Dataset_Results}, CVAE-USM again dominates, with accuracy consistently above 70\% and peaking at 82.12\% in certain scenarios. This dominance indicates CVAE-USM's strong adaptability and learning capabilities across different conditions in the PAMAP2 dataset. TrC, SOT and DANN show respectable performances in specific scenarios, suggesting that they may be better adapted to situations with minimal variance between training and testing data or where data patterns are more consistent and less complex. In contrast, CORAL struggles significantly, with their accuracies mainly ranging between 30\%-40\%. This poor performance reflects their limited adaptability to the PAMAP2 dataset. A critical aspect contributing to these varied performances is the utilization of temporal relationships within the data. CVAE-USM's effective use of temporal information likely contributes to its outstanding accuracy, as it can capture and learn from the dynamics and temporal patterns inherent in the PAMAP2 dataset. In contrast, models like SOT and DANN do not leverage temporal data, which could explain their not outstanding results.

Overall, across both the OPPT and PAMAP2 datasets, CVAE-USM consistently achieves the highest accuracy, proving its adaptability, robustness, and effectiveness in handling cross-user HAR tasks. TrC shows a reasonable performance across both datasets, indicating its potential adaptability to different scenarios, albeit not at the level of CVAE-USM. CORAL, SOT, and DANN exhibit varied and mixed results, suggesting that their effectiveness and adaptability are conditional based on specific dataset characteristics and scenarios. Some methods, particularly CORAL, show significant struggles in adapting to these datasets, highlighting the need for further development to enhance their effectiveness in time series data that involves temporal relation knowledge.

\subsection{Effect of temporal information}

In this section, we analyze the impact of integrating temporal relation knowledge into cross-user HAR, concentrating on the OPPT and PAMAP2 datasets. The comparison involves SOT, a traditional domain adaptation method; DANN, known for its deep domain adaptation approach; and our CVAE-USM, which is specifically designed for time series domain adaptation.

\begin{figure}[h!]
\centering
\includegraphics[width=\columnwidth]{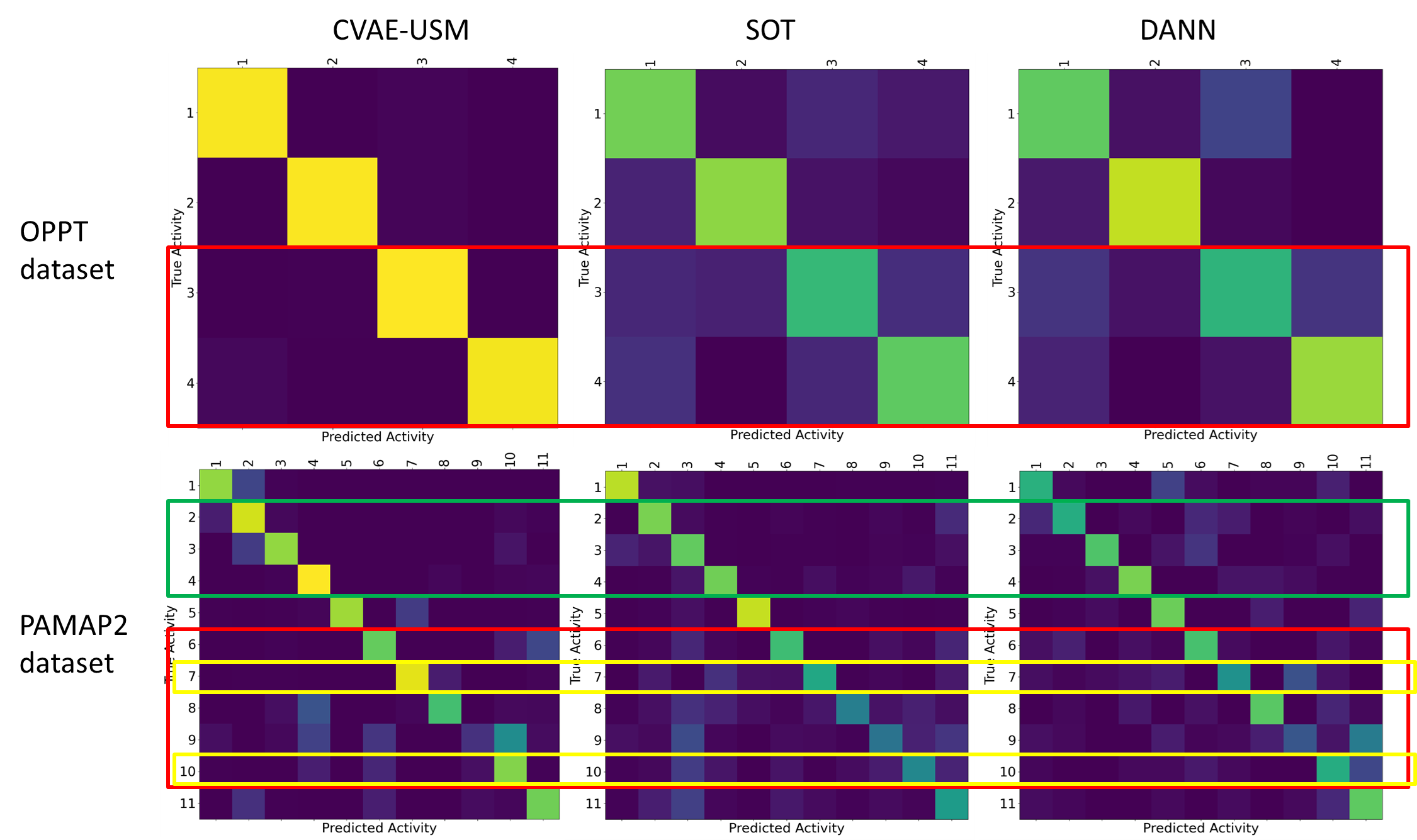}
\caption{Confusion matrices of CVAE-USM, SOT, and DANN methods for the average performance of tasks in the OPPT and PAMAP2 datasets.\label{cm_results}}
\end{figure}

Figure~\ref{cm_results} displays the confusion matrices for CVAE-USM, SOT, and DANN, reflecting their average performance across the OPPT and PAMAP2 datasets. The matrices use color gradients, with warmer colors (closer to yellow) indicating higher values. The x-axis represents predicted activities, while the y-axis corresponds to the actual activities. The diagonal line from the top left to the bottom right of these matrices signifies correct predictions, where the predicted activity aligns with the true activity.

In these analyses, all methods exhibit pronounced values along their diagonals, indicating accurate recognition of most activities and underscoring their effectiveness in cross-user scenarios. Common activities such as 'lying', 'walking', 'standing' and 'cycling' are efficiently identified by all methods. The reason could be attributed to their distinct movement patterns, making them easily distinguishable from other activities. However, confusion arises in differentiating activities with inherent similarities, such as activities like 'ascending stairs' and 'descending stairs'.

CVAE-USM, in particular, excels in identifying activities with complex temporal relations like 'Nordic walking' and 'vacuum cleaning' as the yellow squares shown in Figure~\ref{cm_results}, with higher accuracy compared to SOT and DANN. Additionally, while SOT and DANN show dispersed classification outcomes, CVAE-USM exhibits a more focused and clear classification pattern due to its effective handling of temporal states and extraction of sub-activities as the red squares shown in Figure~\ref{cm_results}. CVAE-USM demonstrates a superior ability to distinguish activities like 'standing', 'sitting', and 'walking' as the green squares shown in Figure~\ref{cm_results}, possibly due to its nuanced capture of temporal relation context and subtle sensor data variations. Moreover, transitional movements within activities, often overlooked by SOT and DANN, are effectively captured by CVAE-USM. For example, repositioning a vacuum cleaner during 'vacuum cleaning' is better understood and classified by CVAE-USM, thanks to its comprehension of temporal relations.

In conclusion, the use of temporal relation knowledge in CVAE-USM effectively reduces the variance in activity patterns between different users, especially in dynamic activities with complex temporal relations. Extracting temporal relation knowledge from time series data enhances the robustness and accuracy of activity recognition. This approach is especially beneficial for activities with rhythmic or sequential traits, allowing for a more nuanced understanding of various activities and highlighting the importance of temporal knowledge in improving cross-user activity recognition capabilities.

\section{Conclusion}
In this study, we present a new approach called CVAE-USM for cross-user HAR tasks from time series data. This method is particularly effective in adapting to different users' data by focusing on the temporal relation knowledge of activities. CVAE-USM uses VAE generative model and adversarial learning techniques to align different users' data distributions, leading to more accurate activity recognition.

The validation is implemented on two public datasets show that CVAE-USM outperforms existing methods in recognizing activities across various users. This success demonstrates CVAE-USM's potential to analyse time series data in HAR. Overall, CVAE-USM represents its advantages in understanding temporal relations and adapting to the complex activity data from different users. In future, we aim to explore the application of CVAE-USM to more complex or less structured activities, to evaluate its adaptability and robustness. 

 \bibliographystyle{named} 
 \bibliography{ref}

\end{document}